\documentclass{article}

\usepackage[final]{neurips_2021}

\usepackage{natbib}
\setcitestyle{numbers,comma,square}

\usepackage[utf8]{inputenc} 
\usepackage[T1]{fontenc}    
\usepackage{hyperref}       
\usepackage{url}            
\usepackage{booktabs}       
\usepackage{amsfonts}       
\usepackage{nicefrac}       
\usepackage{microtype}      
\usepackage{xcolor}         

\usepackage{wrapfig}
\usepackage{graphicx}
\usepackage{url} 

\title{No News is Good News: A Critique of the One Billion Word Benchmark}

\author{
    Helen Ngo\thanks{Correspondence to: Helen Ngo \texttt{<helen@cohere.ai>}, Nicholas Frosst \texttt{<nick@cohere.ai>}}\thanks{Cohere, Toronto, Canada.}
    \\\And João G.M. Araújo\footnotemark[2]
    \\\And Jeffrey Hui\footnotemark[2]
    \\\And Nicholas Frosst\footnotemark[1]\footnotemark[2]
}

\begin{document}

\maketitle

\begin{abstract}

  The One Billion Word Benchmark is a dataset derived from the WMT 2011 News Crawl, commonly used to measure language modeling ability in natural language processing. We train models solely on Common Crawl web scrapes partitioned by year, and demonstrate that they perform worse on this task over time due to distributional shift. Analysis of this corpus reveals that it contains several examples of harmful text, as well as outdated references to current events.
  We suggest that the temporal nature of news and its distribution shift over time makes it poorly suited for measuring language modeling ability, and discuss potential impact and considerations for researchers building language models and evaluation datasets.
  
\end{abstract}

\section{Introduction}

\begin{wrapfigure}{r}[0.01cm]{0.6\textwidth}
\includegraphics[width=7.7cm]{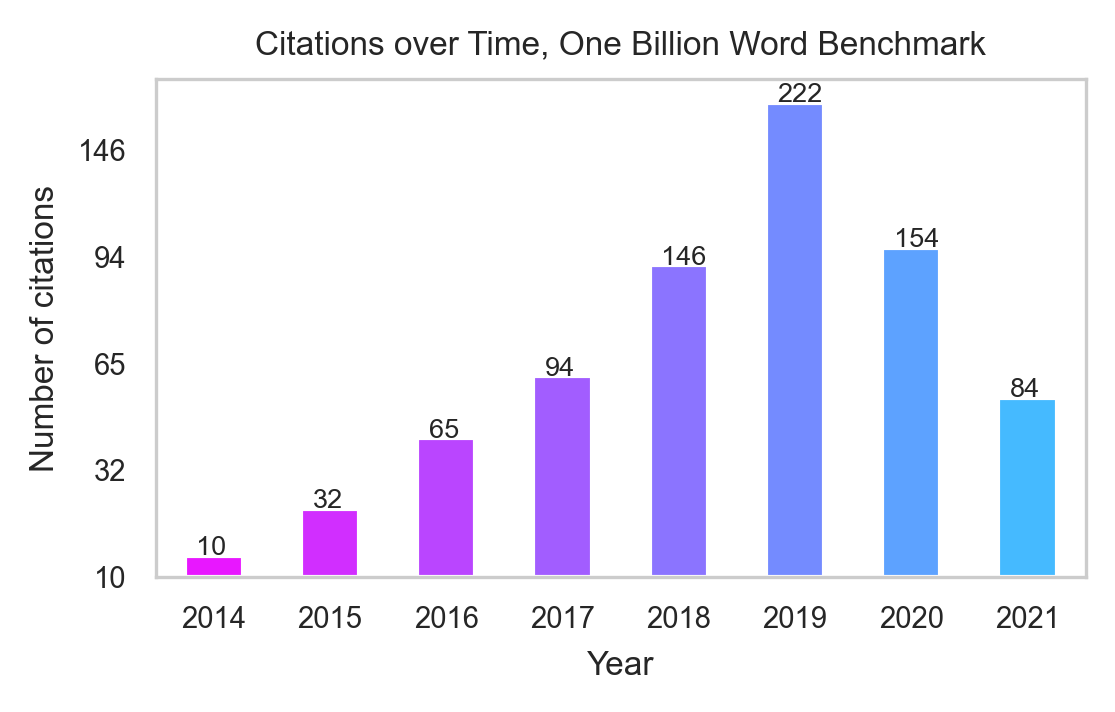}
\centering
\caption{Citations for \texttt{lm1b} are decreasing, but it is still widely used. Source: Semantic Scholar, September 2021.}
\label{semanticscholar}
\end{wrapfigure}

Language models are commonly evaluated on the One Billion Word Benchmark (\texttt{lm1b}) \citep{lm1b}, reporting performance on perplexity \citep{radford2019language}. \texttt{lm1b} is derived from the WMT 2011 News Crawl dataset\footnote{\url{http://www.statmt.org/wmt11/translation-task.html}} released ten years ago. Though the number of citations per year is decreasing, it is still widely used by researchers in recent years, as shown in Figure \ref{semanticscholar}. Examples in the dataset were constructed by extracting single sentences from news articles, such as:  
\begin{itemize}
  \item \texttt{At least 101 people were killed in the blasts.}
  \item \texttt{Who can possbly ever take her serious?}
  \item \texttt{Well, Dexter caught up with me.}
\end{itemize}

Previous work has documented examples of decontextualized hate speech \citep{ngo2021mitigating} within this dataset, as well as the impact of its destructive preprocessing \citep{radford2019language}. Our work suggests that the prevalence of this dataset in the literature is concerning, as evaluation on this dataset inadvertently incentivizes the creation of language models which optimize for generating language in the style of news articles without any regard for factuality, and encodes world knowledge which becomes progressively more outdated. We demonstrate that model performance on this task decreases over time due to distributional shift of the training data, and argue that news articles should not be used as the basis for assessing language modeling ability.


\section{Making generative language models useful in the real world}

Autoregressive language models such as GPT-3 are trained to predict the next token in a sequence given the previous tokens \citep{brown2020language}. Language modeling ability is often reported in perplexity. Perplexity can be viewed as a measure of uncertainty when predicting the next token in a sequence, where the perplexity of a language model for a learned distribution $Q$ and an empirical distribution $P$ can be defined as $Perplexity(P, Q) = e^{crossentropy(P, Q)}$. Using \texttt{lm1b} to measure language modeling ability by reporting perplexity can be viewed as evaluating a model's ability to generate sentences from the same distribution as news articles which were scraped prior to 2011. At the time of its publication, \citep{lm1b} was one of the largest language modeling datasets made publicly available and fully reproducible.

Researchers and policymakers alike are concerned about the risk of malicious actors weaponizing language models for automated disinformation; specifically, generating targeted propaganda in form of fake news articles which closely mimic the language and style of real news \citep{zellers2020defending}. 
Fake news contributes to the erosion of democracy, justice, and public trust \citep{zhou2019}, with several methods being developed for its detection \citep{Shu2017FakeND, zellers2020defending}. News websites also vary in their factuality and media bias \citep{nakov2021survey}. Indiscriminately scraping for news without assessing media bias or factuality results in datasets which capture and reflect inherent media bias.

\citep{benderparrot} note that language models are not grounded in any model of the world, and instead produce sequences in the same linguistic form as observed in its training data without any mechanism to account for factuality. As a result, models trained and evaluated on news corpora such as \texttt{lm1b} will generate text in the linguistic style of news, without any grounding in the real world.

In addition to potential harms from models which are inadvertently optimized for generating fake news, \texttt{lm1b} contains sentences which contain words commonly found on blocklists \citep{ngo2021mitigating}. While these sentences may have plausibly been used in expository contexts within the article, the destructive sentence-level preprocessing and shuffling applied to \texttt{lm1b} removes all long-range structure from the text \citep{radford2019language} and makes it infeasible to track the context and intent of individual examples. \citep{ngo2021mitigating} cross-reference the held-out split of \texttt{lm1b} with an existing blocklist and surface several examples of sentences which contain commonly blocklisted words. As models are evaluated on perplexity, which can be viewed as a proxy for the model's ability to correctly predict the next token, evaluating a language model on this corpus incentivizes the development of models which are better at generating toxic text. For these reasons, while training and testing on \texttt{lm1b} can be used to assess model capability to learn a specific fixed distribution, \texttt{lm1b} evaluation performance should not be reported as an indicator of language modeling ability or as a proxy for the practical utility of a language model.

\section{The limitations of temporal data}

Language models pretrained on text corpora from the open web will not capture knowledge about the world from outside the temporal window of the corpus curation process. For example, GPT-3 was trained on scrapes of Common Crawl\footnote{https://commoncrawl.org/the-data/} from 2016 to 2019 \citep{brown2020language}, and cannot reliably generate information about current events which happened after the 2019 cutoff window. 
\texttt{lm1b} was released in 2013 using a web scrape from 2011, which means that models evaluated on \texttt{lm1b} today are attempting to represent world knowledge about events which happened at least a decade ago. These events no longer represent the world as it exists today, and measuring language models on their ability to generate text from news articles from 2011 inherently penalizes models which learn more recent, up-to-date world representations. 

\begin{figure}[h]
\includegraphics[width=9cm]{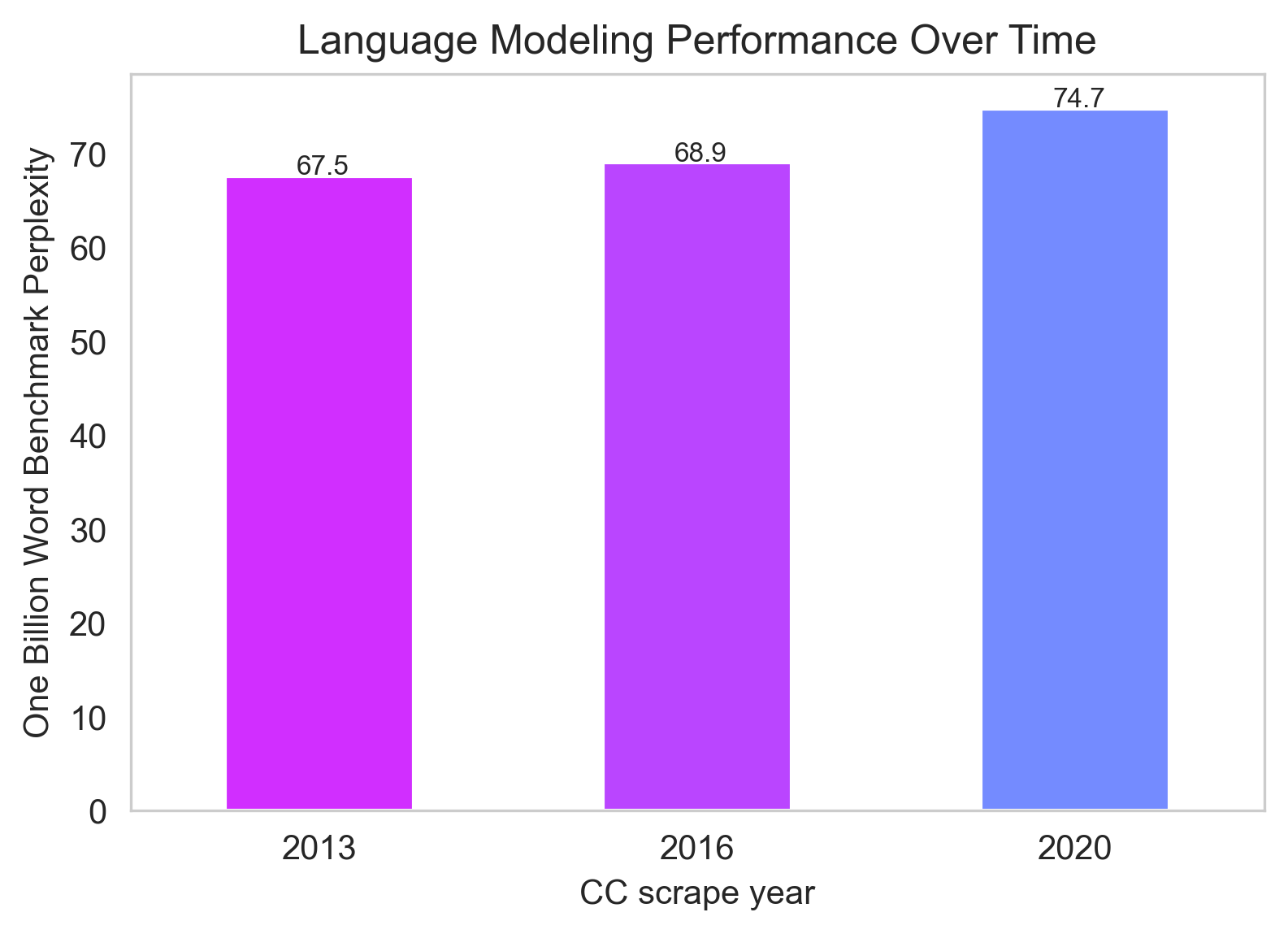}
\centering
\caption{Models trained on Common Crawl scrapes perform worse on the One Billion Word Benchmark over time.}
\label{cctime}
\end{figure}

Common Crawl is a repository of web scrapes of the internet updated annually and is often used as a key data source for language models built on the open web \citep{radford2019language,brown2020language,benderparrot}. We train benchmark models on three distinct datasets created by selecting data sampled from different years of Common Crawl: 2013 (the year which \texttt{lm1b} was released), 2016, and 2020. Each benchmark model followed a standard decoder-based Transformer architecture \citep{radford2019language,vaswani2017attention} with 128M parameters, and was trained on a 10 GB randomly-sampled subset of Common Crawl for 70k steps on Cloud TPU v2 chips using the Adam optimizer with initial learning rate = 0.0001, batch size = 1024, sequence length = 1024, embedding dimension = 1024, layers = 6, attention heads = 4. Figure \ref{cctime} demonstrates the negative impact of distribution shift over time. Models which are trained on datasets temporally further removed from the \texttt{lm1b} corpus source (i.e. WMT 2011 News Crawl dataset) exhibit higher perplexity than those trained on datasets which are temporally closer.

\section{Discussion}

\citep{rogers2021changing} makes the case that researchers building language models should be purposeful in curating training datasets, as curation choices are effectively world design choices. Corpora built on top of news scrapes snapshotted at a specific point in time will capture all of the inherent social bias and structural issues related to news reporting at a given point in time. \citep{kaplan2020scaling} demonstrate that models with fixed sizes will be capacity-limited, further highlighting the need for careful data curation: in practice, most deployed language models are fixed-size, and care must be taken to ensure that they are learning from the highest-quality data possible to make the best use of their capacity.

While it can be argued that all data scraped from the internet is representative of a specific snapshot of the cultural zeitgeist, news in particular should be processed within a specific cultural context. \citep{Castillo_2014} demonstrate that engagement around a particular news article decays rapidly after its initial posting. News which is relevant within a particular moment in culture may not be relevant to the public consciousness weeks or even days from its initial posting, and resources should not be maximized for optimizing language models which model outdated or irrelevant news, current events, and cultural context.

Synthetic news generation also does not broadly appear to be a relevant or useful goal for researchers working on language modeling, and the threat of fake news powered by language models undermines the integrity of our information ecosystem. 

For these reasons, we encourage researchers developing language modeling benchmarks to actively avoid building evaluation datasets based on snapshots of news corpora from snapshots fixed in time. Potential future distribution shift and inherent media bias make them unsuitable for evaluating language models built on an ever-evolving open web.

\section{Conclusion}
We demonstrate that language models trained on corpora which is temporally further removed from the distribution of \texttt{lm1b} perform worse on \texttt{lm1b} over time due to distributional shift. We also outline shortcomings of \texttt{lm1b} and news data in general as it pertains to utility as a language modeling benchmark, and argue that  \texttt{lm1b} should be only be used to assess model capability to learn a specific fixed distribution across training and test data, as opposed to an indicator of general language modeling capability.


\section*{Acknowledgements}
We thank Cooper Raterink for his helpful feedback on the manuscript. This experiments in this work were made possible because of the  infrastructure built by the Cohere team.

\newpage 


\end{document}